\documentclass[conference]{IEEEtran}
\IEEEoverridecommandlockouts
\usepackage{cite,algorithm,algorithmic,graphicx,textcomp,xcolor}
\usepackage{amsmath,amssymb,amsfonts}
\usepackage{bm,tabularx,url}
\newcolumntype{L}[1]{>{\raggedright\let\newline\\\arraybackslash}m{#1}}
\newcolumntype{C}[1]{>{\centering\let\newline\\\arraybackslash}m{#1}}
\newcolumntype{R}[1]{>{\raggedleft\let\newline\\\arraybackslash}m{#1}}
\def\BibTeX{{\rm B\kern-.05em{\sc i\kern-.025em b}\kern-.08em
    T\kern-.1667em\lower.7ex\hbox{E}\kern-.125emX}}

\graphicspath{{figures/}}
\begin{document}
\title{Learn a Prior for RHEA \\for Better Online Planning\\
\thanks{\textsuperscript{*} Corresponding author.}
\thanks{Code and data are available at https://github.com/for-xintong/p-RHEA.}
}

\author{\IEEEauthorblockN{Xin Tong}
\IEEEauthorblockA{\textit{School of Information Science} \\
\textit{and Technology, USTC}\\
Hefei, China \\
txt@mail.ustc.edu.cn}
\and
\IEEEauthorblockN{Weiming Liu}
\IEEEauthorblockA{\textit{School of Information Science} \\
\textit{and Technology, USTC}\\
Hefei, China \\
weiming@mail.ustc.edu.cn}
\and
\IEEEauthorblockN{Bin Li \textsuperscript{*}}
\IEEEauthorblockA{\textit{School of Information Science} \\
\textit{and Technology, USTC}\\
Hefei, China \\
binli@ustc.edu.cn}
}

\maketitle

\begin{abstract}

Rolling Horizon Evolutionary Algorithms (RHEA) are a class of online planning methods for real-time game playing; their performance is closely related to the planning horizon and the search time allowed. In this paper, we propose to learn a prior for RHEA in an offline manner by training a value network and a policy network. The value network is used to reduce the planning horizon by providing an estimation of future rewards, and the policy network is used to initialize the population, which helps to narrow down the search scope. The proposed algorithm, named prior-based RHEA (p-RHEA), trains policy and value networks by performing planning and learning iteratively. In the planning stage, the horizon-limited search assisted with the policy network and value network is performed to improve the policies and collect training samples. In the learning stage, the policy network and value network are trained with the collected samples to learn better prior knowledge. Experimental results on OpenAI Gym MuJoCo tasks show that the performance of the proposed p-RHEA is significantly improved compared to that of RHEA.

\end{abstract}


\section{Introduction}

MCTS \cite{Chaslot:2006,Browne:2012} has been widely used in domains of sequential decision making such as game playing. It first builds a game tree with game states as nodes and actions as edges, and then uses a forward model to simulate actions before making a final decision about the move to take. In continuous game domains, Rolling Horizon Evolutionary Algorithms (RHEA) \cite{Perez:2013, Perez:2015} have been viewed as a suitable alternative to MCTS. RHEA approaches encode a sequence of actions into an individual, then adopt Evolutionary Algorithms (EA) \cite{Eiben:2003} to optimize the action sequences directly. The agent uses EA to evolve the population during some predefined budget, then selects the first action of the best individual as the move to take in the real game. 

The performance of RHEA is closely related to its planning horizon $H$. A longer planning horizon allows RHEA to plan from a more global perspective, while the scale of search space grows exponentially with the extension of the planning horizon. On the other hand, a longer planning horizon asks the EA to deal with a higher dimensional optimization problem, which is still a critical issue and active research area in the EA community \cite{Li:2013}. A shorter planning horizon allows RHEA to search more quickly, but it considers only the short-term rewards, which probably weaken the agent's performance when future rewards are inconsistent with short-term rewards. 

To tackle the above dilemma, in this paper, an algorithm named prior-based RHEA (p-RHEA) is proposed to enhance RHEA in two ways. On the one hand, a value network is introduced to estimate the future rewards after $H$ steps to provide a long-term view with limited planning horizon. On the other hand, a policy network is trained to initialize the population for RHEA, which can narrow down the search to high-probability actions and save the search time. These two neural networks are trained by performing planning and learning in an iterative way. In each cycle, p-RHEA performs planning in simulated games with the prior knowledge provided by policy and value networks, while the samples generated in search are collected to train the networks in order to learn better prior knowledge. By iteratively performing planning and learning, p-RHEA continuously gets better samples and better neural networks. 

The rest of this paper is organized as follows: Section \ref{Related} briefly introduces the existing methods for games. Section \ref{Background} introduces the background closely related to our work. Section \ref{Method} shows the details of the proposed p-RHEA algorithm, and section \ref{Experimental} gives the experimental setup and results analysis. Finally, further discussions are presented in Section \ref{Conclusions}.

\section{Related Works}\label{Related}

\subsection{Model-based planning methods}

Model-based planning approaches do not need training but require a forward model to allow the agent to revisit the states it has experienced. MCTS \cite{Chaslot:2006} is a typical model-based planning method. In each simulation iteration, it performs four steps: selection, expansion, evaluation and backup. A tree policy, such as UCT \cite{Kocsis:2006}, is used to select actions with higher action values plus a bonus to encourage exploration. When the iterations have finished, the action with the highest average reward or the maximum number of visits will be executed. RHEA \cite{Perez:2013} optimizes a series of fixed-length action sequences with corresponding cumulative rewards as their fitness, then selects the first action of the best individual as the move to take. The value of the last state is evaluated by running a rollout \cite{Tesauro:1997} to the end of the game with a fast rollout policy or simply by a heuristic \cite{Perez:2015}. Other than these, model predictive control (MPC) \cite{Garcia:1989, Tassa:2014} has a long history in robotics and control systems. It represents the next state as a function of the current state and controller and then uses the gradient information to optimize the controller to maximize the cumulative reward. Therefore, an analytical expression about the rewards is additionally needed. 

\subsection{Model-free learning methods}

Model-free learning approaches need training but do not require a forward model. The agent always goes from a reset state to a terminal state, then starts over. Many reinforcement learning (RL) methods fall into this category. DQN \cite{Mnih:2015} is a value-based RL method; it aims to approximate the optimal action-value function to make decisions. In contrast to value-based methods, policy-based RL methods such as A3C \cite{Mnih:2016} and PPO \cite{Schulman:2017} directly parameterize the policy and update the parameters by performing gradient ascent on the expected value of the total cumulative reward. There are also gradient-free methods to optimize the parameters of the policy network, such as CEM \cite{Szita:2006} and NES \cite{Ruckstiess:2010}. These population-based approaches often face the problem called "curse of dimensionality", because each individual encodes a deep neural network. ERL \cite{Khadka:2018} and CEM-RL \cite{Pourchot:2018} combine gradient-free and gradient-based methods for policy search. They are hybrid algorithms that leverage the population of an EA to provide diversified samples to train an RL agent, and reinsert the RL agent into the EA population periodically to inject gradient information into the EA. 

\subsection{Combined methods}

Combined approaches here refer to methods that combine online planning and offline learning, thus require both a forward model and training. They learn from the samples generated by model-based planning, and the knowledge gained can, in turn, be used for a better online search. AlphaGo \cite{Silver:2016} and AlphaGo Zero \cite{Silver:2017} use a neural network to guide the search of MCTS in the game of Go. The neural network improves the strength of tree search, resulting in higher quality move selection. POLO \cite{Lowrey:2018} combines local trajectory optimization with global value function learning and explains how approximate value function can help reduce the planning horizon and allow for better policies beyond local solutions. 

\section{Background}\label{Background}

\subsection{Rolling Horizon Evolutionary Algorithms}

Here we introduce the main ideal of RHEA approaches and the implementation details. In each state $s$, RHEA can be viewed as facing a trajectory optimization problem: 
\begin{equation}
\label{equ:RHEA}
\begin{aligned}
\hat{\pi}(s)=\mathop{\arg\max}_{a_{0:H-1}|s_0=s}\mathbb{E}\bigg[\sum_{t=0}^{H-1}\gamma^t r(s_t,a_t)+\gamma^H r_f(s_H)\bigg]
\end{aligned}
\end{equation}
where $r(s_t,a_t)$ represents the reward obtained by performing action $a_t$ under state $s_t$ and $r_f(s_H)$ represents a final reward function. $r_f(s_H)$ can be estimated by performing rollout multiple times, but is often omitted due to time constraints. $\gamma \in (0, 1)$ is the discount factor. A smaller $\gamma$ means greater weights are given to the more recent rewards. We encode a sequence of actions into an individual and then use a specific EA to optimize (\ref{equ:RHEA}). Fitness values are calculated by executing the sequence of actions of the individual using the forward model, until all actions are executed or a terminal game state is reached. As the sequence is optimized, denoted by $a_{0:H-1}^*$, the first action $a_0^*$ is executed, and the procedure is repeated. In the initialization stage of the next cycle, the first $H-1$ actions of the first individual can be initialized with $a_{1:H-1}^*$, which can make full use of the results of the previous cycle to guide the search. 

In continuous control domains, each action $a$ is a real vector, so the dimension of the problem to be optimized is equal to the planning horizon multiplied by the action dimension. Another point worth noting is that RHEA is an open loop approach, as the states found during the search are not stored or reused in any way \cite{Perez:2015}. 

\subsection{Policy evaluation and policy improvement}
Here we introduce some basic concepts of policy iteration \cite{Sutton:2018}, which are necessary for our p-RHEA. We first introduce a policy $\pi(s)$, it inputs the current state $s$ and outputs the action that should be taken under state $s$. Then we introduce a value function $V_\pi(s)$ that estimates how good the state $s$ is under policy $\pi(s)$. It should be noted that a value function is defined with respect to a particular policy. The value of state $s$ under policy $\pi(s)$ is the average discounted reward accumulated by following the policy from the state: 
\begin{equation}
\label{equ:value_function}
\begin{aligned}
V_\pi(s)=\mathbb{E}\bigg[\sum_{t=0}^{\infty}\gamma^t r(s_t,\pi(s_t)) \,\big|\, s_0=s\bigg]
\end{aligned}
\end{equation}

The initial policy and value function are randomly initialized. Then given the policy, we can update its value function in any state $s$, which is called policy evaluation: 
\begin{equation}
\label{equ:policy_evaluation}
\begin{aligned}
V_\pi^{new}(s)=\mathbb{E}\bigg[r(s,\pi(s)) + \gamma V_\pi(s^{'})\bigg]
\end{aligned}
\end{equation}
where $s^{'}$ is the state encountered after state $s$ under policy $\pi(s)$. The purpose of computing the value function for a policy is to help find a better policy:
\begin{equation}
\label{equ:policy_improvement}
\begin{aligned}
\pi^{new}(s)=\mathop{\arg\max}_{a}\mathbb{E}\bigg[r(s,a)+\gamma V_\pi(s^{'})\bigg]
\end{aligned}
\end{equation}

This process of getting a new policy that improves on an original policy, by making it greedy with respect to the value function of the original policy, is called policy improvement. Once a policy $\pi(s)$ has been improved using $V_\pi(s)$ to yield a better policy $\pi^{new}(s)$, we can then compute $V_\pi^{new}(s)$ and use it again to yield an even better policy. We can thus obtain a sequence of improved policies and value functions. This way of finding an optimal policy is called policy iteration. 

\section{Method}\label{Method}

As mentioned before, the proposed p-RHEA method has two neural networks to learn and store prior knowledge compared to the conventional RHEA approaches. The policy network $p(a|s;\theta)$ takes the current state as input and outputs the probability distribution of the action that should be taken under the state. The value network $V_\pi(s;\theta_v)$ also takes the current state as input but outputs a scalar which represents the future cumulative reward from the state. We use $\theta$ and $\theta_v$ to represent the parameters of the policy network and value network, respectively. These two neural networks are initialized to random weights, and we do not share parameters between policy and value networks. The proposed p-RHEA method consists of two stages: training (Section \ref{training}) and real play (Section \ref{planning}). 

\subsection{Online planning with the learned prior knowledge}\label{planning}

Given the policy network and value network, we can run p-RHEA online, similar to the RHEA process. The objective of online planning is shown as follow:
\begin{equation}
\label{equ:pRHEA_improvement}
\begin{aligned}
\pi(s)=\mathop{\arg\max}_{a_{0:H-1}|s_0=s}\mathbb{E}\bigg[\sum_{t=0}^{H-1}\gamma^t r(s_t,a_t)+\gamma^H V_\pi(s_H;\theta_v)\bigg]
\end{aligned}
\end{equation}
where $s$ is the current state and $a_{0:H-1}$ is the action sequence p-RHEA trying to optimize. After optimization, the first action $a_0$ will be performed at current state,  then p-RHEA continues to the next cycle. The online planning process of p-RHEA is concluded below:
\begin{enumerate}
	\item Initialize the population using the prior probability provided by the policy network. Specifically, the current state $s_0$ is first fed into the policy network to obtain a probability distribution, then an action $a_0$ is sampled from the distribution and executed to reach the next state $s_1$. After $H$ steps, a sequence of actions is sampled, which can be encoded into an individual. This process is repeated for $NP$ (population size) times to get a population. Compared to random initialization, the initial population of p-RHEA is guided in a more promising region, thus saving many search costs. 
	\item Evolve the population with a specific EA. Individuals are evaluated by performing the sequence of actions in the simulated environment. The fitness of an individual is defined as $R = \sum_{t=0}^{H-1}\gamma^t r(s_t,a_t)+\gamma^H V_\pi(s_H;\theta_v)$. Future rewards after $H$ steps are estimated by the value network. With the help of the value network, p-RHEA can plan from a more global perspective and skip some local optimum while limiting the planning horizon. 
	\item When the evolution is complete, we record the optimal individual $a_{0:H}^*$ and its fitness $R^*$. Then we take a step $a^*=a_0^*$, and the game is simulated to the next state. 
	\item Repeat the above steps until a terminal state is encountered, which indicates the end of the game. 
\end{enumerate}

We use CMA-ES \cite{Hansen:2016} for action sequence optimization, which utilizes a multivariate Gaussian distribution to model the correlation between variables and is proved to perform well on problems with tens of dimensions \cite{Omidvar:2014}. A multivariate Gaussian model is chosen because actions need to work in coordination, which means we are dealing with a nonseparable optimization problem. 

\subsection{Offline training to learn better prior knowledge}\label{training}

\begin{figure}[!h]
	\centering
	\includegraphics[width=0.9\columnwidth]{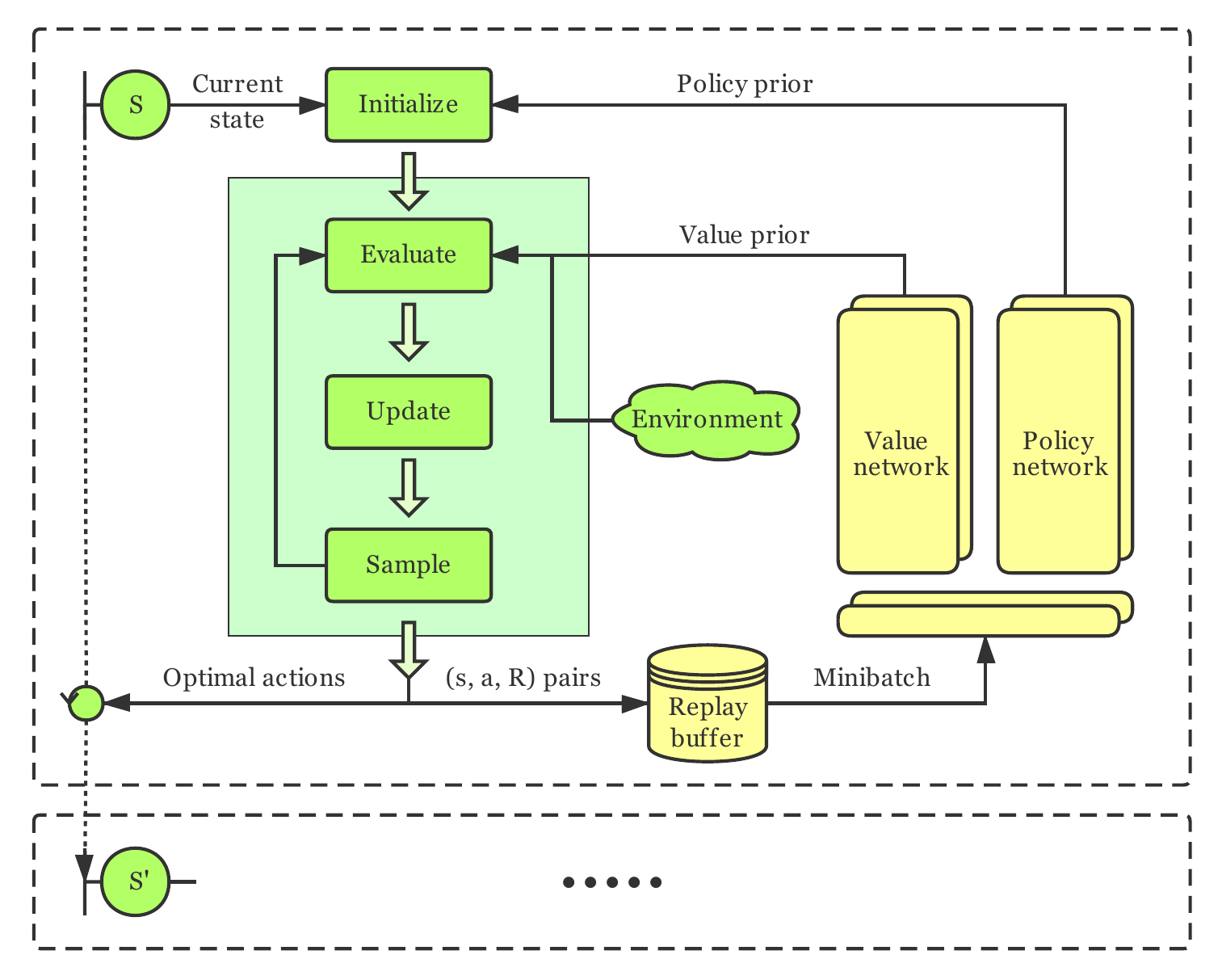}
	\caption{High level training schematic of p-RHEA, highlighting the combination of planning (green) and learning (yellow)}
	\label{fig:framework}
\end{figure}

The training flow of the proposed p-RHEA approach is shown in Fig. \ref{fig:framework}, it performs planning and learning in an iterative way. In each simulation cycle, planning is first performed with the prior knowledge provided by policy and value networks. The sample pair $(s, a^*, R^*)$ generated by the search is stored into a replay buffer $D$ for training, with old experiences discarded if the buffer becomes full. The experience replay mechanism used here can remove correlations in the sample sequence \cite{Mnih:2015}. From the perspective of RL, lookahead search can be similarly considered to be both a policy improvement operator and a policy evaluation operator: After the planning stage, we get an improved policy for state $s$ and a new value estimate of state $s$ under policy $\pi(s)$. 
\begin{equation}
\label{equ:pRHEA_iteration}
\begin{aligned}
\pi^{new}(s) = a^*,\qquad V_\pi^{new}(s) = R^*
\end{aligned}
\end{equation}

After every $T$ steps of planning in the game and collecting experience, the policy network and value network are updated by learning from minibatches sampled uniformly from the replay buffer. In the learning stage, the parameters of the policy network $\theta$ are adjusted to maximize the similarity of the policy network action probability $p(a|s;\theta)$ to the optimal action $a^*$. The parameters of the value network $\theta_v$ are adjusted to minimize the error between the new value estimate $R^*$ and the old value estimate $V_\pi(s;\theta_v)$. Specifically, the neural networks are trained by RMSProp optimizer \cite{LeCun:2015} on a loss function that sums over the maximum likelihood loss and mean squared error, respectively: 
\begin{equation}
	\label{equ:loss}
	\begin{aligned}
		Loss = -\log p(a^*|s;\theta) + \frac{1}{2}(R^*-V_\pi(s;\theta_v))^2
	\end{aligned}
\end{equation}

The offline training of p-RHEA is a closed loop process. In the planning stage, the prior knowledge provided by the policy network and value network is used to achieve faster and more global search. While the sample pairs generated by search can in turn be used to train the networks in the learning stage to provide better prior knowledge. Through such iterations, we continually get better samples and better prior knowledge, and finally we can expect far better performance than the beginning. 

p-RHEA requires additional training stage compared to RHEA, but once training is completed, p-RHEA is expected to achieve better performance with less search cost and shorter planning horizon in real play stage. Two tricks are used when training our p-RHEA.  First, the value prior is added after a certain number of samples (more than or equal to $V_{start}$) have been collected, which ensures that the initial search will not be biased due to random initialization of the value network. A randomly initialized policy network does not introduce bias into the search and is effective from the beginning. Second, as the sequence $a_{0:H-1}^*$ is optimized, not only the first action but the first $\lfloor H/2\rfloor$ actions will be executed. Correspondingly, the first $\lfloor H/2\rfloor$ sample pairs will be added to the replay buffer. One can refer to Algorithm \ref{algo:p-RHEA} in the Appendix for more details. This change only occurs during the training stage, with the aim of speeding up the collection of samples and reducing the time required for training. It will make p-RHEA inferior on the training curve, but good prior knowledge can still be learned for real play, as shown in the next section. 

\section{Experimental Studies and Results}\label{Experimental}

In order to test the performance of our approach, we evaluate p-RHEA and comparison algorithms in several continuous control tasks simulated with the MuJoCo physics engine \cite{Todorov:2012}. The states of the simulated robots are their generalized positions and velocities, and the controls are joint torques. These MuJoCo tasks have one to seventeen joints to control and allow the agent to take up to 1000 steps. Every step of the agent, the environment will simulate to the next state and return a reward. The analytical expression of the reward is unknown, but the agent is allowed to revisit the states it has experienced. 

\subsection{Performance of RHEA with Different Horizons}

We first test the performance of RHEA on the MuJoCo tasks and investigate the impact of planning horizon $H$ on the final performance. CMA-ES is selected for trajectory optimization and the population size ($NP$) is set to $4+\lfloor 3\log(D)\rfloor$ as suggested. $D$ is the dimension of the problem to be optimized, which is equal to planning horizon multiplied by the number of joints of the simulated robot. Two different planning horizons are implemented, $H=20$ and $H=50$. The former is optimized for 20 generations, and the latter is optimized for 50 generations. The discount factor $\gamma$ is set to 0.99. The results of the means and standard deviations of 5 independent runs are listed in Table \ref{tab:RHEA}.

\begin{table}[!h]
	\caption{Performance of RHEA with different horizons on MuJoCo tasks}
	\begin{center}
		\begin{tabular}{|C{0.19\textwidth}|C{0.1\textwidth}|C{0.1\textwidth}|}
			\hline
			& RHEA H=20 & RHEA H=50 \\
			\hline
			Ant-v2 & $4891\pm75$ & $2681\pm299$ \\
			\hline
			HalfCheetah-v2 & $6857\pm379$ & $17586\pm1481$ \\
			\hline
			Hopper-v2 & $285\pm14$ & $439\pm82$ \\
			\hline
			Humanoid-v2 & $716\pm113$ & $2135\pm1064$ \\
			\hline
			InvertedPendulum-v2 & $321\pm257$ & $1000\pm0$ \\
			\hline
			InvertedDoublePendulum-v2 & $607\pm118$ & $3056\pm1828$ \\
			\hline
			Swimmer-v2 & $43\pm2$ & $52\pm4$ \\
			\hline
			Walker2d-v2 & $182\pm39$ & $1089\pm560$ \\
			\hline
		\end{tabular}
		\label{tab:RHEA}
		\vspace*{-10pt}
	\end{center}
\end{table}

We can see that the results of looking ahead 50 steps are generally better than those of looking ahead 20 steps, except on the Ant-v2 task. As mentioned before, a longer planning horizon allows RHEA to plan from a more global perspective and is expected for better performance. But for Ant-v2 task, it has eight joints to control. When looking ahead 50 steps, CMA-ES has to deal with a 400-dimensional problem which is quite difficult to optimize. Maybe the trajectories found are not as good as those look ahead 20 steps. Without any prior knowledge, RHEA has achieved the best results on the Ant-v2 and HalfCheetah-v2 tasks as far as we know. On these two tasks, RHEA chooses a novel way forward, not running or jumping but rolling, which allows the simulated robot to have a faster forward speed. We found that RHEA works very well on tasks that do not have deceptive rewards. However, RHEA almost failed on the Hopper-v2 and Walker2d-v2 tasks. These two tasks are typical environments where there are deceptive rewards: In order to get more short-term rewards, the simulated robot will lean forward as much as possible. But once it falls, which is defined by thresholds on the torso height and angle, the game is over, and no future rewards will be given any more. A video demonstration of how RHEA acts on the HalfCheetah-v2 and Hopper-v2 tasks is available at \url{https://github.com/for-xintong/p-RHEA-video1}.

In summary, RHEA needs a long planning horizon for better performance, but this requires more search costs. In addition to this, RHEA lacks global information to guide it to identify deceptive rewards. Next, we will show how a pre-learned knowledge can help solve the above problems. 

\subsection{Performance of p-RHEA}

The p-RHEA approach is implemented using CMA-ES as well, but only looks ahead 20 steps and optimizes for 5 generations. To represent the policy, we use a fully-connected multilayer perceptron (MLP) with two hidden layers of 128 units and ReLU nonlinearities to output the mean of univariate Gaussian distribution. The log-standard deviation is parameterized by a global vector independent of the state, as done in \cite{Schulman:2015, Duan:2016}. The value network is also constructed with two hidden layers, each with 128 units, to output a scalar which represents the future cumulative reward. 

\begin{figure}[!h]
	\centering
	\begin{minipage}[t]{0.48\linewidth}
		\centering
		\includegraphics[width=1.1\textwidth]{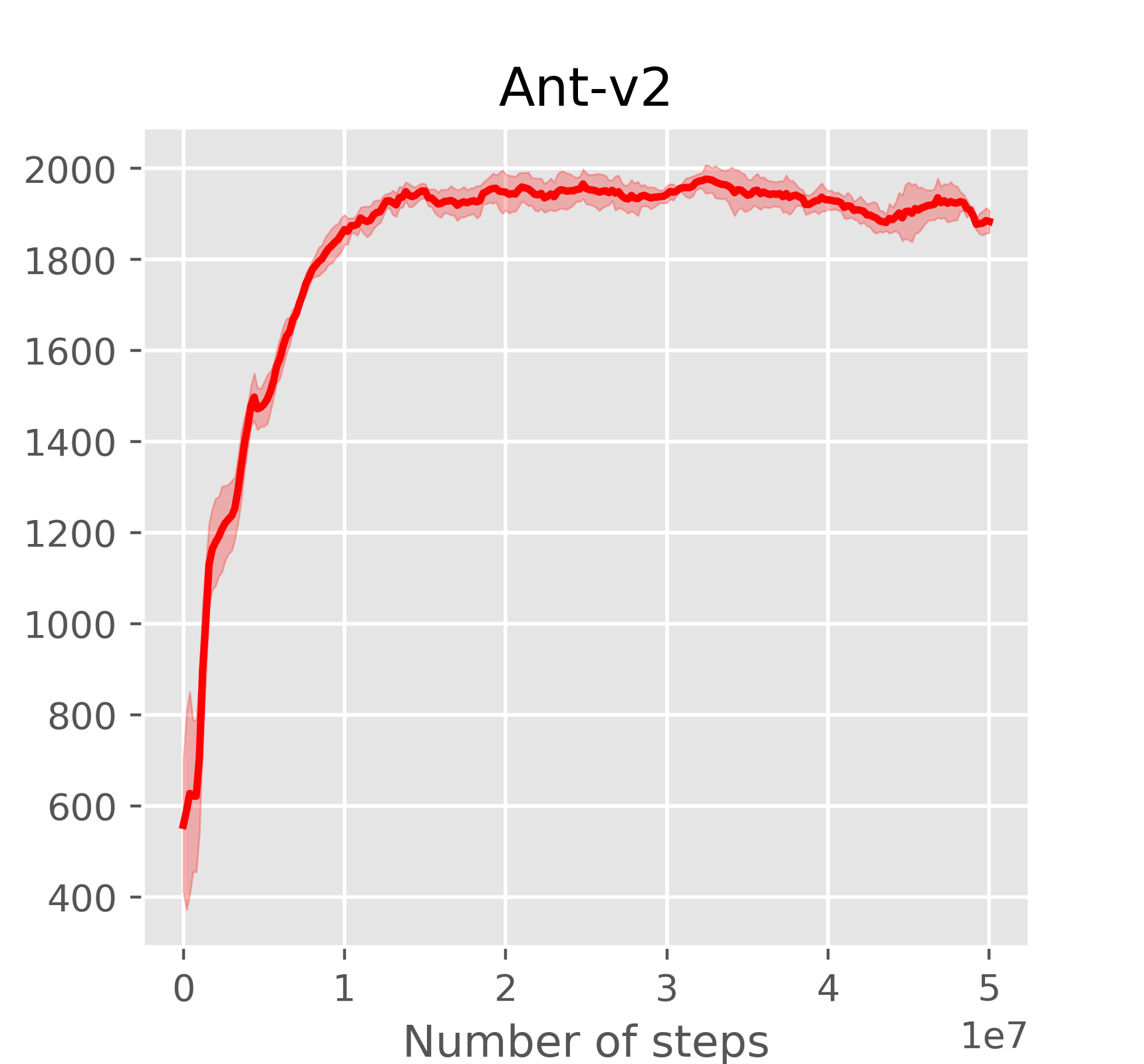}
	\end{minipage}	
	\begin{minipage}[t]{0.48\linewidth}
		\centering
		\includegraphics[width=1.1\textwidth]{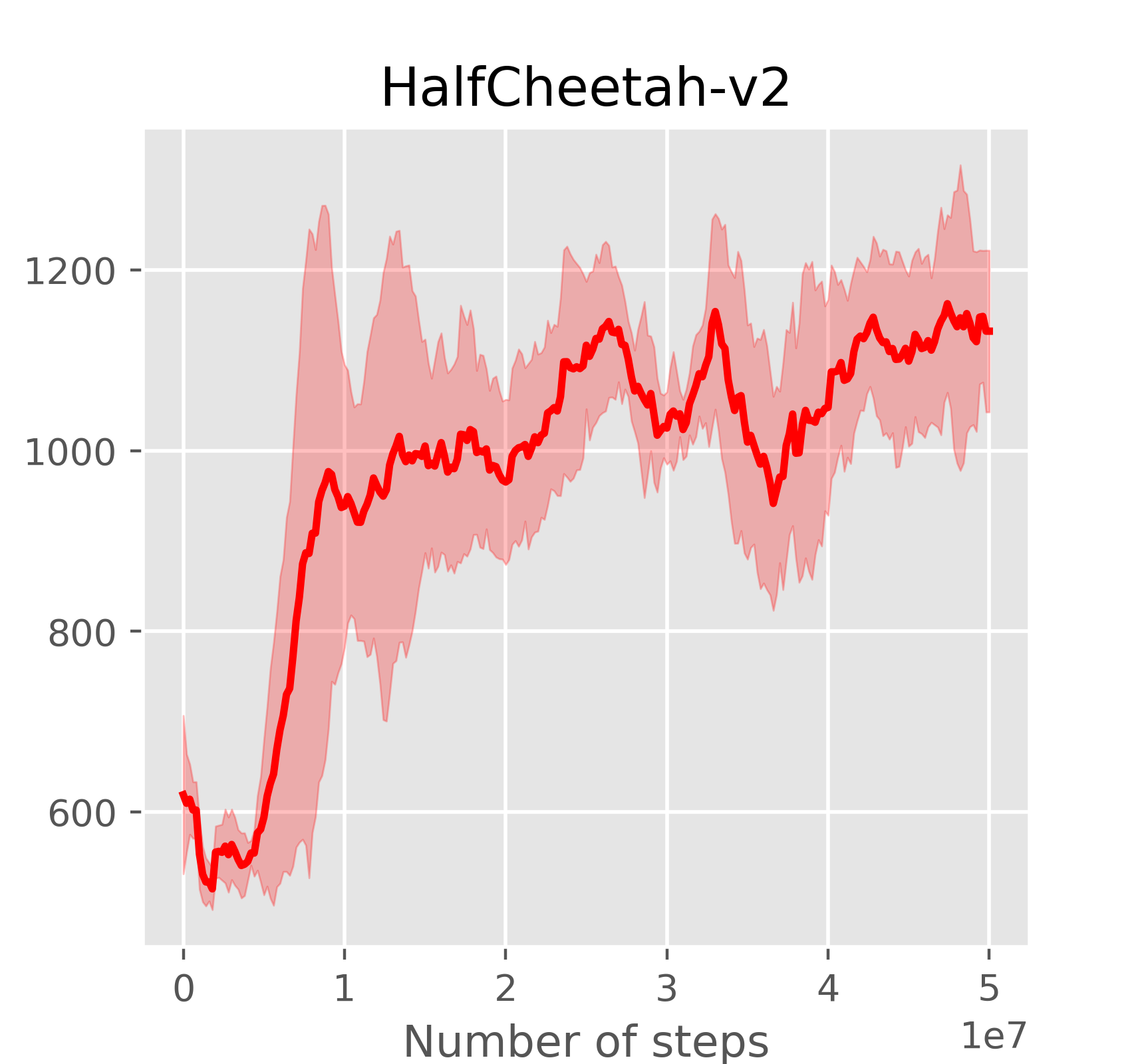}
	\end{minipage}
	\begin{minipage}[t]{0.48\linewidth}
		\centering
		\includegraphics[width=1.1\textwidth]{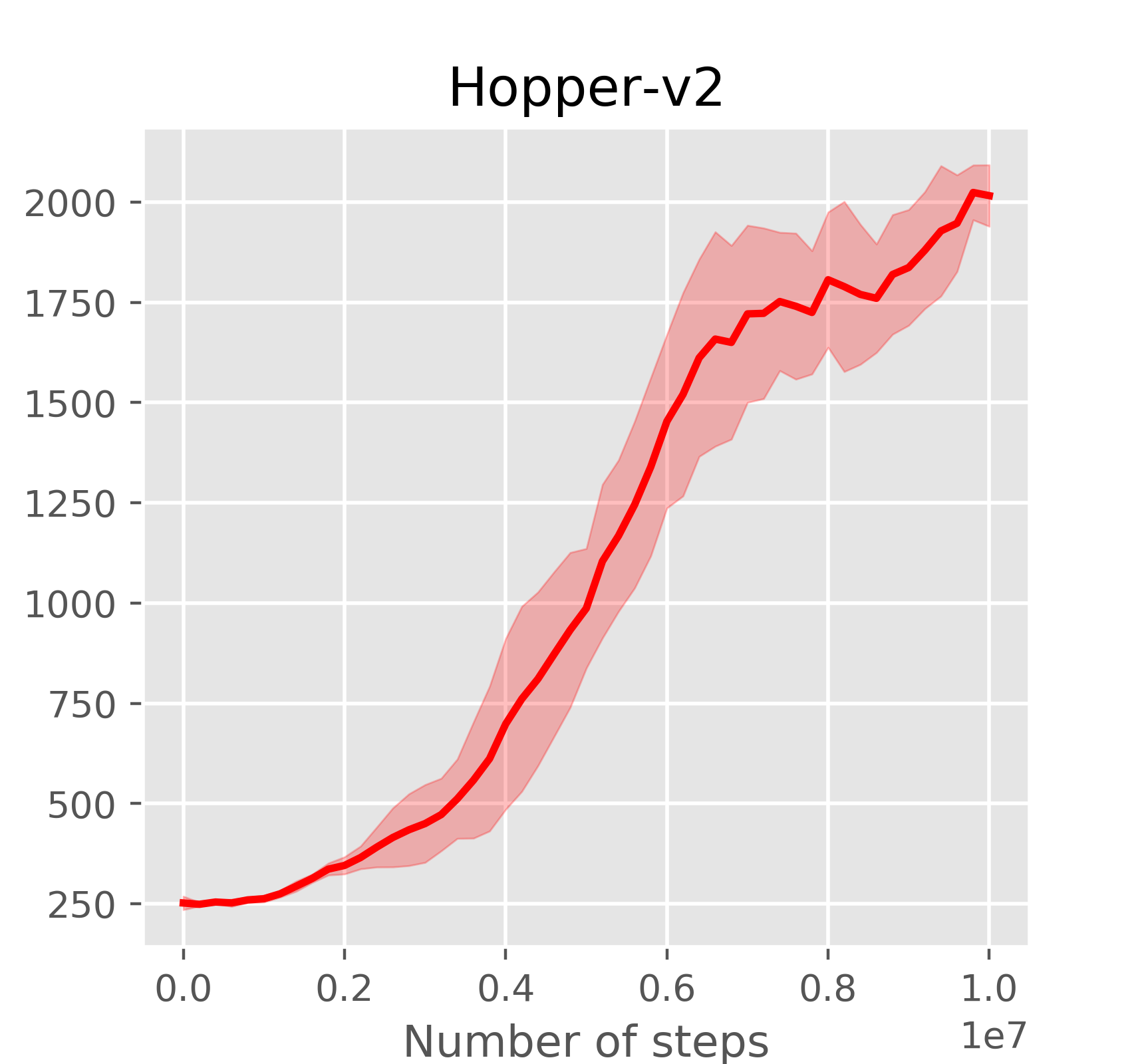}
	\end{minipage}	
	\begin{minipage}[t]{0.48\linewidth}
		\centering
		\includegraphics[width=1.1\textwidth]{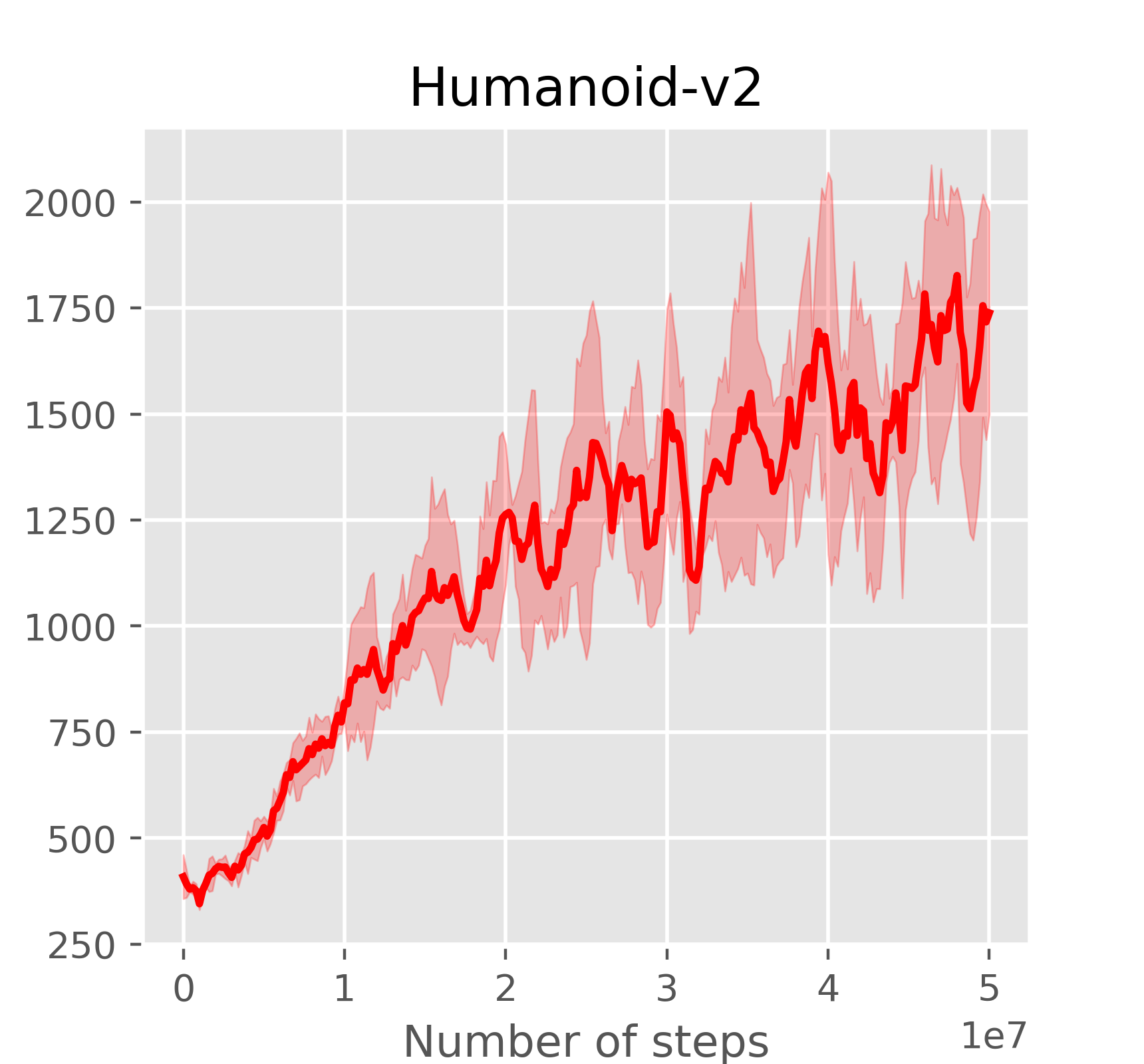}
	\end{minipage}
		\begin{minipage}[t]{0.48\linewidth}
		\centering
		\includegraphics[width=1.1\textwidth]{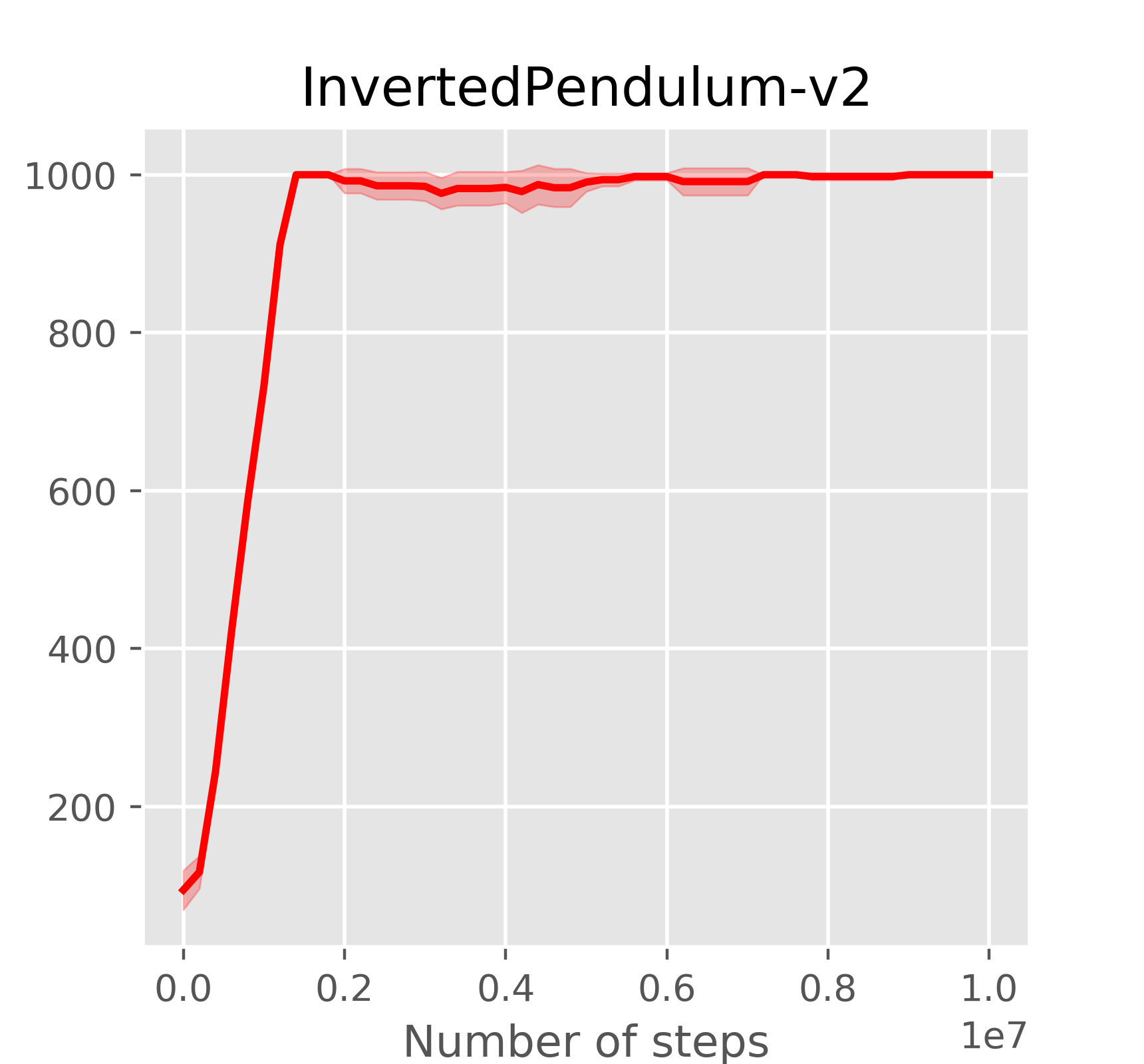}
	\end{minipage}
	\begin{minipage}[t]{0.48\linewidth}
		\centering
		\includegraphics[width=1.1\textwidth]{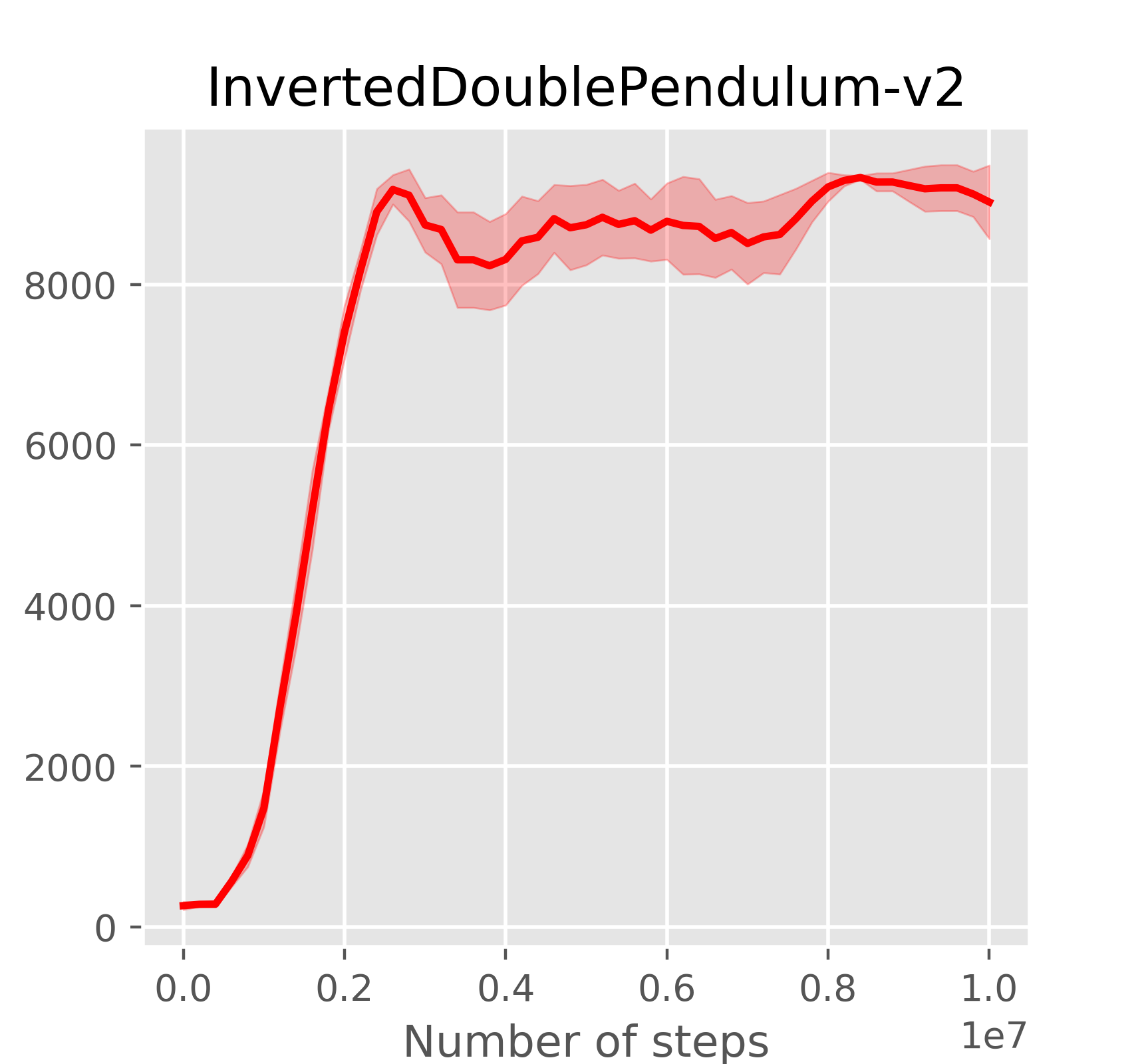}
	\end{minipage}
	\begin{minipage}[t]{0.48\linewidth}
		\centering
		\includegraphics[width=1.1\textwidth]{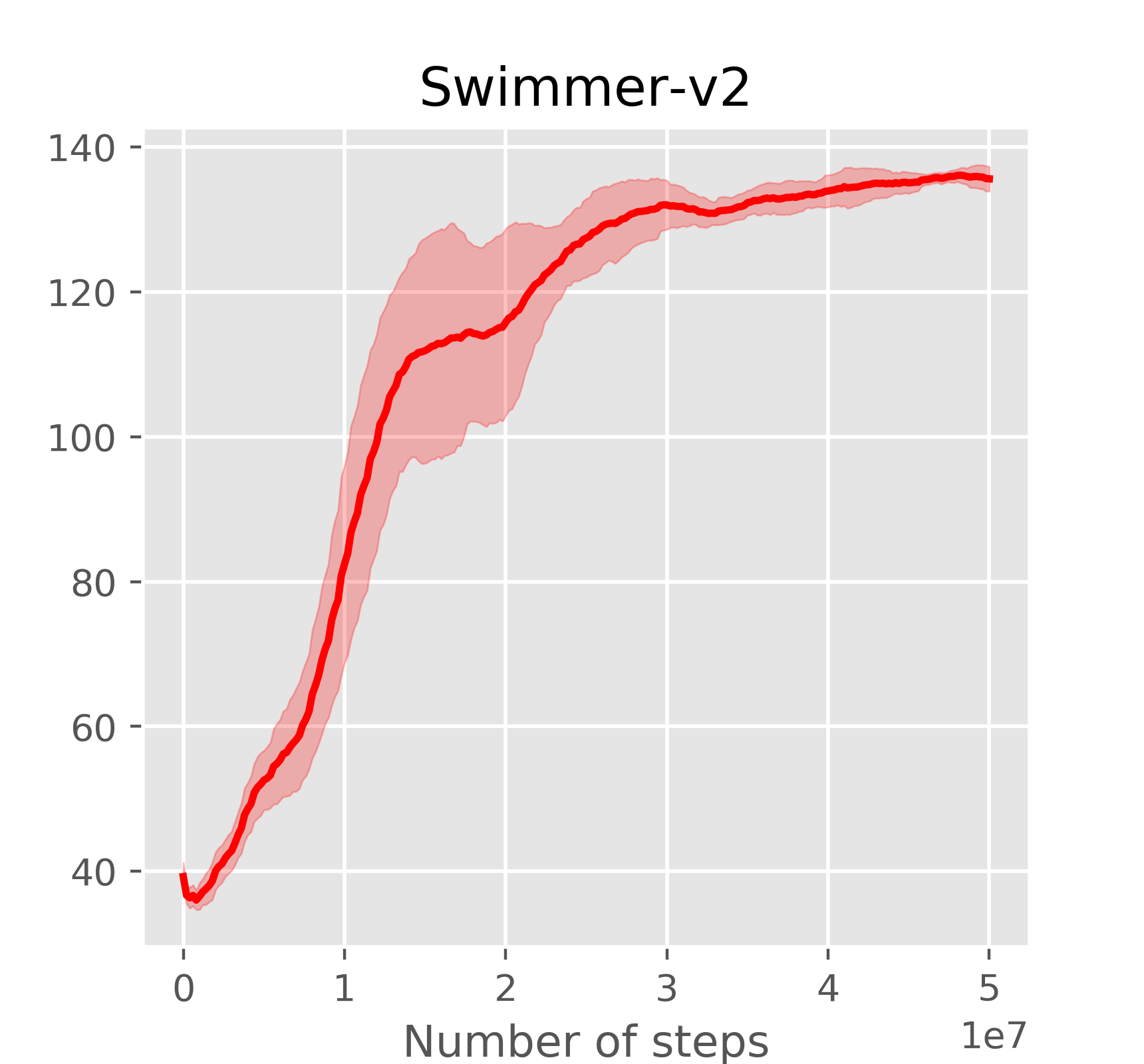}
	\end{minipage}
	\begin{minipage}[t]{0.48\linewidth}
		\centering
		\includegraphics[width=1.1\textwidth]{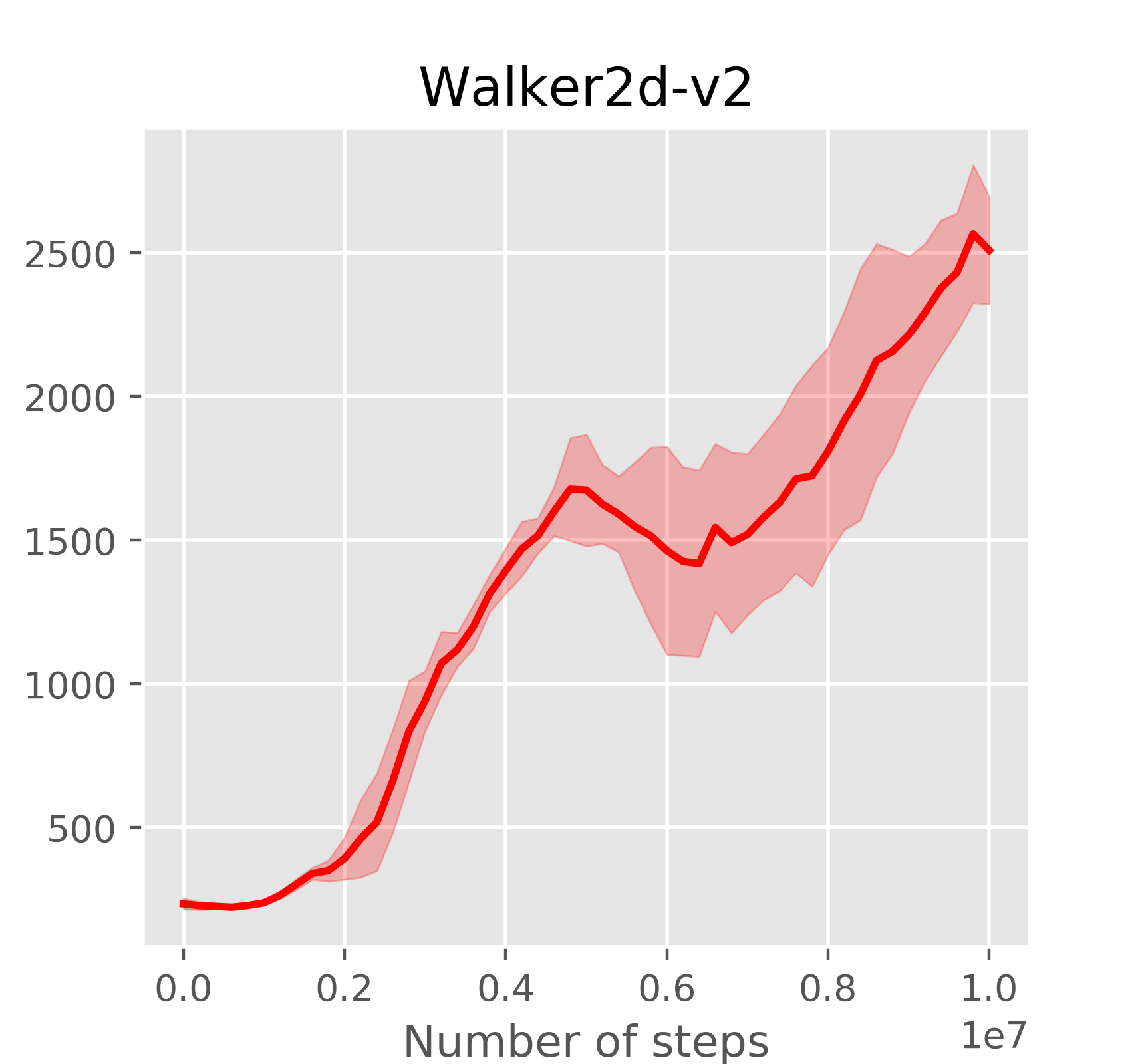}
	\end{minipage}
	\caption{Training curves of p-RHEA on the MuJoCo tasks, the means and standard deviations are calculated from 5 independent runs}
	\label{fig:curve}
	\vspace*{-20pt}
\end{figure}

\begin{table*}[!h]
	\caption{Performance of p-RHEA and two comparison algorithms on MuJoCo tasks}
	\begin{center}
		\begin{tabular}{|C{0.19\textwidth}|C{0.1\textwidth}|C{0.1\textwidth}|C{0.1\textwidth}|}
			\hline
			& RHEA H=50 & PPO2 & p-RHEA H=20 \\
			\hline
			Ant-v2 & $2681\pm299$ & $897\pm315$ & $\bm{3229}\pm\bm{202}$ \\
			\hline
			HalfCheetah-v2 & $\bm{17586}\pm\bm{1481}$ & $1669\pm438$ & $3162\pm411$ \\
			\hline
			Hopper-v2 & $439\pm82$ & $2316\pm796$ & $\bm{2795}\pm\bm{727}$ \\
			\hline
			Humanoid-v2 & $2135\pm1064$ & $615\pm110$ & $\bm{4531}\pm\bm{878}$ \\
			\hline
			InvertedPendulum-v2 & $\bm{1000}\pm\bm{0}$ & $809\pm88$ & $\bm{1000}\pm\bm{0}$ \\
			\hline
			InvertedDoublePendulum-v2 & $3056\pm1828$ & $7103\pm1567$ & $\bm{9344}\pm\bm{3}$ \\
			\hline
			Swimmer-v2 & $52\pm4$ & $111\pm3$ & $\bm{138}\pm\bm{4}$ \\
			\hline
			Walker2d-v2 & $1089\pm560$ & $3425\pm768$ & $\bm{3901}\pm\bm{639}$ \\
			\hline
		\end{tabular}
		\label{tab:p-RHEA}
		\vspace*{-10pt}
	\end{center}
\end{table*}

The means and standard deviations of training curves on MuJoCo tasks are shown in Fig. \ref{fig:curve}. The results are from 5 independent runs. The total number of training steps is set to 10M, which is generally set to 1M in the field of model-free reinforcement learning. It should be pointed out that the samples put into the replay buffer is only about 0.05M. Like AlphaGo Zero which takes 1600 simulations to perform a move, our p-RHEA takes about 200 simulations to find a good action and use it for training. p-RHEA has low sampling efficiency but good sample quality because each sample is carefully selected. Ant-v2, HalfCheetah-v2 and Swimmer-v2 tasks are given more training steps because each of their episodes takes 1000 steps (no early termination will occur) and 1M training steps only allow them to play about 50 episodes. In addition, Humanoid-v2 is considered a hard task and is also given more training steps. 

We can see that the score of p-RHEA is getting higher and higher as the training progresses, which means useful prior knowledge is being learned. For the HalfCheetah-v2 and Swimmer-v2 tasks, the scores of the training curves have a small drop when the value network is first involved (at the very beginning of the curves) which means that a randomly initialized value network can have bad guidance for search. However, through training, the value network (along with the policy network) will help achieve better performance as expected. Each task takes several hours to train on a 4-core CPU computer. After the training stage is completed, we test the performance of p-RHEA for real play. At this time, p-RHEA optimizes a 20-step trajectory and only takes the first action to execute. The means and standard deviations of 25 runs with the learned knowledge are listed in Table \ref{tab:p-RHEA}. For comparison, the results of the PPO method, which taken from OpenAI Baselines \cite{Baselines:2017}, are also listed in the table. 

In general, p-RHEA is significantly better than RHEA and PPO approaches. Compared to model-free reinforcement learning methods, p-RHEA can benefit from a forward model, which allows it to perform multiple simulations and then take the best action to step. For example, PPO performs poorly on the Ant-v2 and Humanoid-v2 tasks which have eight and seventeen joints to control, respectively. Through online planning, p-RHEA can achieve much better results than just learning a policy offline. Compared to model-based online planning methods, p-RHEA can benefit from the value prior and policy prior, which allow it to plan from a more global perspective and with less search cost. For example, without the global information provided by policy and value networks, the simulated robot controlled by RHEA will soon fall due to seeking for high short-term rewards on the Hopper-v2 task and get trapped in local optimum due to avoiding big negative rewards on the Swimmer-v2 task. In contrast, p-RHEA performs well on these two tasks by taking advantage of the prior knowledge learned, a video demonstration is available at \url{https://github.com/for-xintong/p-RHEA-video2}. On the HalfCheetah-v2 task, however, p-RHEA can be considered as a failure compared to RHEA. We checked it carefully and found it is because we look ahead 20 steps and then take the first ten steps instead of the first one during training. This trick, as mentioned before, can reduce the time required for training. Taking ten steps per cycle performs well on other tasks, but on the HalfCheetah-v2 task, it makes the simulated robot sometimes fail to turn over because of insufficient speed. A video demonstration of how the number of steps taken can affect the performance on the HalfCheetah-v2 task is available at \url{https://github.com/for-xintong/p-RHEA-video3}. 

\subsection{How can planning benefit from the prior}

\begin{figure}[!h]
	\centering
	\begin{minipage}[t]{0.9\linewidth}
		\centering
		\includegraphics[width=1.1\textwidth]{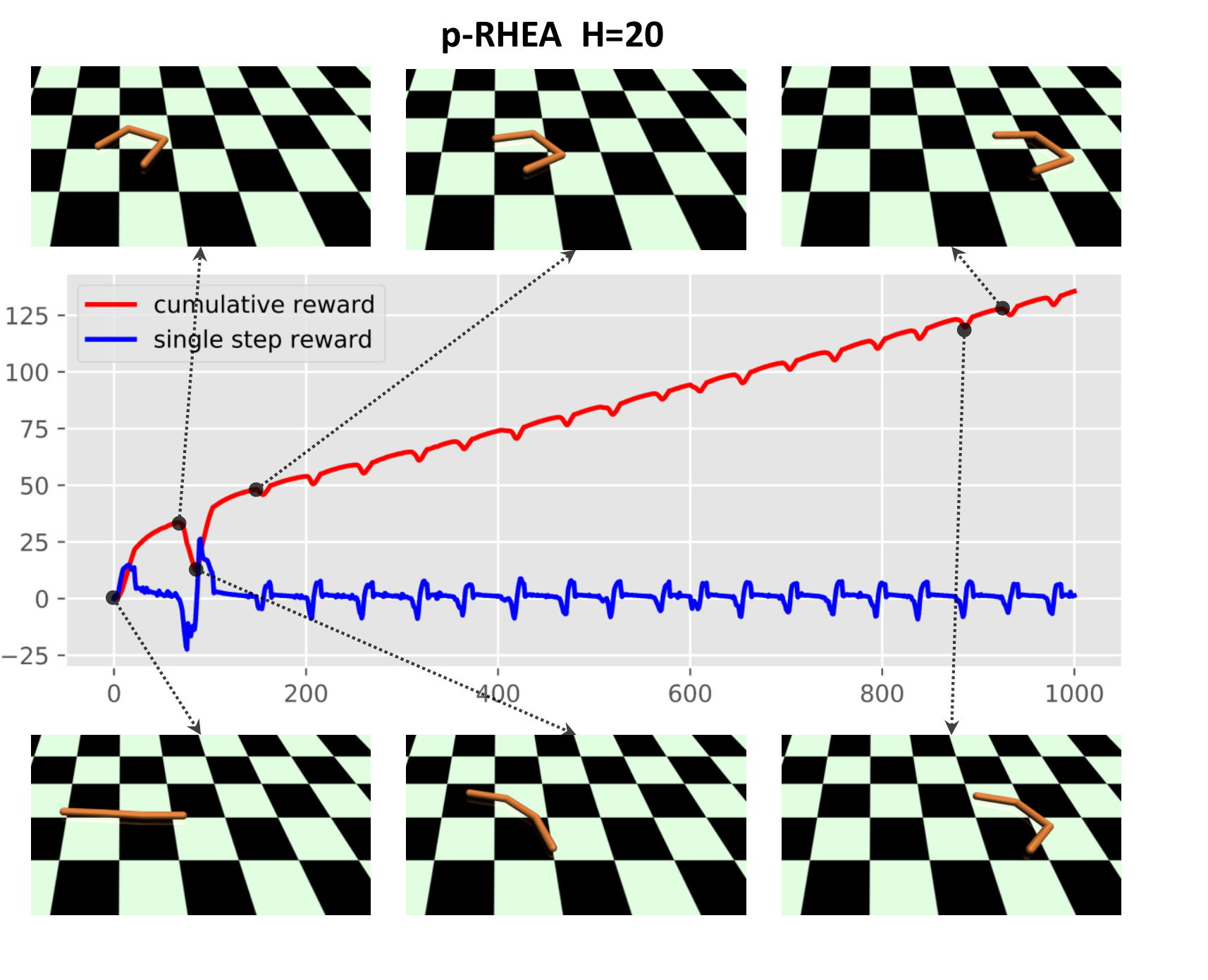}
	\end{minipage}
	\begin{minipage}[t]{0.9\linewidth}
		\centering
		\includegraphics[width=1.1\textwidth]{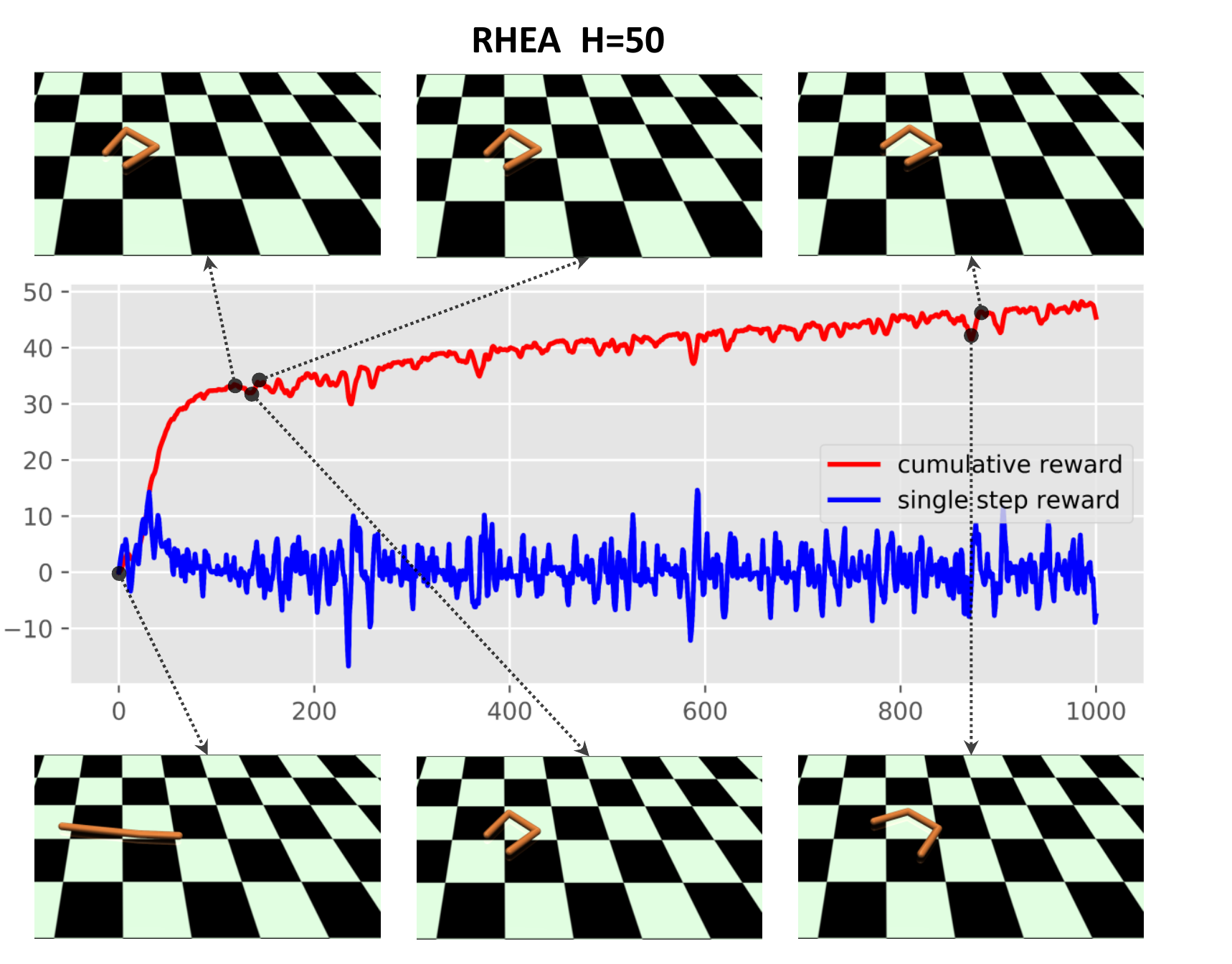}
	\end{minipage}
	\caption{Compare the rewards of p-RHEA and RHEA on Swimmer-v2}
	\label{fig:compare}
	\vspace*{-5pt}
\end{figure}

In order to visualize how the prior knowledge can help p-RHEA make better online planning, we draw the reward curves of the p-RHEA and RHEA methods on the Swimmer-v2 task, as shown in Fig. \ref{fig:compare}. The red lines represent the cumulative reward, and the blue lines represent the single step reward. Each single step reward is multiplied by 10 for better visualization. In addition, we also show the pose of the simulated robot at key time points.

An intuitive feeling is that p-RHEA learned a fixed swimming pattern with the help of prior knowledge. The rewards obtained by p-RHEA are very periodic, but the rewards of RHEA seem to be irregular. For better analysis, we show the reward function of Swimmer-v2, which consists of two parts: a linear reward for forward progress $v_x$ and a quadratic penalty on joint effort $u$. 
\begin{equation}
\label{equ:reward}
\begin{aligned}
r = v_x - 10^{-5}||u||^2
\end{aligned}
\end{equation}

The swimming cycle of p-RHEA can be divided into three stages. (i) Closing the legs, which will make the swimmer accelerate and correspond to big positive rewards. (ii) Floating, which uses inertia to advance and corresponds to small positive rewards. (iii) Opening the legs, which will produce a large joint effect. Due to the low speed, the penalty item plays a major role, and the environment returns big negative rewards. The actions of opening the legs which currently appear to be bad are chosen because the agent is informed by the value network that there will be greater positive rewards in the future. For example, when the cumulative reward is about 33, the swimmer controlled by p-RHEA jumps out of the reward trap with just one swimming cycle. In contrast, the swimmer controlled by RHEA does not open its legs significantly, so it is trapped in the local optimum, and the final swimming distance is very short. A video demonstration of how p-RHEA and RHEA act on the Swimmer-v2 task is available at \url{https://github.com/for-xintong/p-RHEA-video4}. 
 
\section{Conclusions and Further Discussions}\label{Conclusions}

In this paper, we proposed a new method called p-RHEA for real-time games, which combines the strengths of planning and learning and shows clear advantages in continuous control tasks. In the training stage, p-RHEA iteratively executes planning and learning in simulated games. With such iterations, p-RHEA is constantly learning from its own experience and performing better and better. In the real play stage, p-RHEA uses the learned value prior to make more global planning while with a shorter planning horizon. In addition to this, p-RHEA uses the learned policy prior to narrow down the search to high-probability actions and can save considerable search costs. 

In the experimental part, we first tested the effect of the planning horizon of RHEA on continuous control tasks. The results showed that longer planning horizon can improve the performance generally, but the computational cost required increases rapidly. We observed that RHEA performs poorly on tasks with deceptive rewards, typical tasks like Hopper-v2 and Swimmer-v2. Then we tested our p-RHEA method. Benefiting from the prior knowledge learned, it only looks ahead 20 steps and optimizes for 5 generations, but can achieve better performance than RHEA which looks ahead 50 steps and optimizes for 50 generations. Finally, we visually demonstrated that prior knowledge learned can help p-RHEA plan from a more global perspective and jump out of local optimums on the Swimmer-v2 task. 

Concerning with the future work, we are trying to test the performance of our p-RHEA algorithm in more complex environments like video games and two-player games. In addition, seeking for a better loss function seems to be a quite promising direction. Considering that CMA-ES can return not only the optimal action sequence but also the distribution of these actions, we can try cross-entropy loss for the update of the policy network and add additional constraints to make the policy update smoother. 

\bibliographystyle{IEEEtran}
\bibliography{CEC19}

\section*{Appendix}

\subsection{Hyperparameters}

\begin{table}[!h]
	\caption{Architecture of the networks}
	\begin{center}
		\begin{tabular}{|C{0.22\textwidth}|C{0.18\textwidth}|}
			\hline
			 Policy network & Value network\\
			\hline
			(state dim, 128) & (state dim, 128) \\
			leaky ReLU, $\alpha$=0.2 & leaky ReLU, $\alpha$=0.2 \\
			(128, 128) & (128, 128) \\
			leaky ReLU, $\alpha$=0.2 & leaky ReLU, $\alpha$=0.2 \\
			(128, action dim) + (action dim) & (128, 1) \\
			\hline
		\end{tabular}
		\label{tab:network}
	\end{center}
\end{table}

\begin{table}[!h]
	\caption{p-RHEA hyperparameters used on MuJoCo tasks}
	\begin{center}
		\begin{tabular}{|C{0.22\textwidth}|C{0.18\textwidth}|}
			\hline
			Hyperparameter & Value \\
			\hline
			Planning horizon ($H$) & 20 \\
			Dimension ($D$) & $H$ * action dim \\
			Population size ($NP$) & $4+\lfloor 3\log(D)\rfloor$ \\
			Number of generations ($NG$) & 5 \\
			Steps taken per cycle ($T$) & 1 \\
			Discount factor ($\gamma$) & 0.99 \\
			\hline
			\hline
			Minibatch size & 32 \\
			Replay buffer size ($V$) & 20000 \\
			Replay start size ($V_{start}$) & 5000 \\
			Training times per cycle ($NT$) & 50 \\
			RMSProp learning rate & 3$\times 10^{-3}$ \\
			RMSProp decay factor & 0.99 \\
			Gradient clipping & 0.5 \\
			\hline
		\end{tabular}
		\label{tab:hyperparameter}
	\end{center}
\end{table}

\subsection{Pseudocode of Algorithm}

\begin{algorithm*}[!h]
	\caption{Pseudocode of p-RHEA for each cycle}
	\label{algo:p-RHEA}
	\begin{algorithmic}[1]
		\REQUIRE{Current environment state $s$, Policy network $p(a|s;\theta)$, Value network $V_\pi(s;\theta_v)$, \\
			Optimal action sequence from the last cycle, Planning horizon $H$, Number of generations ($NG$), \\
			Population size ($NP$), Training times ($NT$), Replay buffer $D$, Replay start size $V_{start}$}
		\STATE{Initialize the parameters of CMA-ES}
		\FOR{$i\leftarrow$ 1 to $NG$}
		\IF{$i = 1$}
		\STATE{Initialize $NP$ $H$-steps action sequences using the policy network, the first action sequence \\
			can take the advantage of the optimal actions from the last cycle}
		\STATE{Encode each action sequence into an individual to form a population}
		\ELSE
		\STATE{Use the mean vector and covariance matrix of CMA-ES to generate a population}
		\ENDIF
		\FOR{$j\leftarrow$ 1 to $NP$}
		\STATE{Set environment state, i.e. $s_0 = s$}
		\STATE{Decode individual $j$ into an action sequence, $\{a_0,a_1,...,a_{H-1}\}$}
		\FOR{$t\leftarrow$ 0 to $H-1$}
		\STATE{Interact with the environment: $r_t,s_{t+1} = Env.step(a_t|s_t)$}
		\IF{$s_{t+1}$ is a terminal state}
		\STATE{break}
		\ENDIF
		\ENDFOR
		\STATE{Length of legal action sequence: $L = t+1$}
		\STATE{$R_{L} = \begin{cases}
			\quad\quad 0 \;\quad,  \text{ $s_{L}$ is a terminal state $||$ size of buffer $D$ $< V_{start}$}\\
			V_\pi(s_{L};\theta),  \text{ otherwise}
			\end{cases}$}
		\FOR{$t\leftarrow$ $L-1$ to 0}
		\STATE{$R_t = r_t + \gamma R_{t+1}$}
		\ENDFOR
		\STATE{Use $R_0$ as the fitness of individual $j$}
		\ENDFOR
		\STATE{Select $\mu=\lfloor NP/2 \rfloor $elite individuals to update the parameters of CMA-ES}
		\ENDFOR
		\STATE{Record the optimal state sequence $\{s_0^*,s_1^*,...,s_{L^*}^*\}$, action sequence $\{a_0^*,a_1^*,...,a_{L^*}^*\}$ and reward \\
			sequence $\{R_0^*,R_1^*,...,R_{L^*}^*\}$} found by CMA-ES
		\STATE{Steps taken per cycle $T = 1$ \\
			(for training, set $T = \max\{1, \lfloor L^*/2\rfloor\}$ to speed up the collection of samples)}
		\STATE{Set current environment state $s=s_T^*$}
		\STATE{Put $\{(s_t^*, a_t^*, R_t^*), t=0,1,..,T-1\}$ sample pairs into the replay buffer $D$}
		\FOR{$i\leftarrow$ 1 to $NT$}
		\IF{size of buffer $D$ $\ge V_{start}$}
		\STATE{Randomly sample a minibatch from the replay buffer $D$}
		\STATE{Train the policy network and value network based on the loss function (\ref{equ:loss})}
		\ENDIF
		\ENDFOR
		\RETURN The remaining optimal action sequence $\{a_{T}^*,a_{T+1}^*,...,a_{L^*}^*\}$ for next cycle
	\end{algorithmic}
\end{algorithm*}

\end{document}